\documentclass[conference]{IEEEtran}
\IEEEoverridecommandlockouts
\usepackage[T1]{fontenc}
\usepackage{cite}
\usepackage{amsmath,amssymb,amsfonts}
\usepackage{graphicx}
\usepackage{textcomp}
\usepackage[table]{xcolor}
\usepackage{booktabs}
\usepackage{multirow}
\usepackage{hyperref}
\hypersetup{hidelinks}
\usepackage{xspace}

\definecolor{remmblue}{HTML}{0072B2}
\definecolor{mutedgray}{HTML}{8A929A}
\definecolor{transportrow}{HTML}{FFF0DB}
\definecolor{sequencerow}{HTML}{E8EFFB}
\definecolor{hiddenrow}{HTML}{EDF4E3}
\newcommand{\evalyes}{\textcolor{remmblue}{$\checkmark$}}
\newcommand{\evalno}{\textcolor{mutedgray}{$\times$}}

\newcommand{\method}{REMM\xspace}

\begin{document}

\title{REMM: Revisable Embodied Memory\\ for Manipulation}

\author{\IEEEauthorblockN{Revanth Krishna Senthilkumaran\textsuperscript{1,2,*},
Ruwan Wickramarachchi\textsuperscript{2},
Alessandro Oltramari\textsuperscript{2},
Jonathan Francis\textsuperscript{1,2}}
\IEEEauthorblockA{\textsuperscript{1}Carnegie Mellon University\\
\textsuperscript{2}Bosch Center for AI\\
\thanks{\textsuperscript{*}Work performed during an internship at Bosch Center for AI.}}}
\hypersetup{pdftitle={REMM: Revisable Embodied Memory for Manipulation},
pdfauthor={Revanth Krishna Senthilkumaran, Ruwan Wickramarachchi, Alessandro Oltramari, Jonathan Francis}}

\maketitle

\begin{abstract}

Long-horizon manipulation requires robots to retain task-relevant history while revising conclusions no longer supported by physical evidence. Retaining observations alone does not specify which completed conditions should be withdrawn after a disturbance or which obligations require repair. We introduce \method, Revisable Embodied Memory for Manipulation, a supervisory memory layer for a frozen G0.5 vision-language-action executor. \method separates an append-only evidence history from revisable task state and explicitly links derived relations and completed conditions to their supporting facts. When accepted evidence changes a supporting fact, it withdraws dependent conclusions and schedules targeted recovery at verified skill boundaries without updating the executor. Across 51 physical trials, including operator-assisted runs, \method improves mean recorded task progress over the strongest baseline from 57\% to 91\% in intervention recovery and from 43\% to 70\% in sequence recall. Complementary video diagnostics show improved grounding of retained history into the next instruction, including comparisons with $\pi_{0.5}$-compatible high-level planner adaptations. These results support revisable memory as a practical interface for continued manipulation without policy retraining. Project: \url{https://revanthsenthil.github.io/remm/}.
\end{abstract}

\section{Introduction}
\label{sec:intro}

Long-horizon manipulation requires more than executing a sequence of independent skills. Each action changes the world in which subsequent decisions are made, and information needed later may no longer be directly observable: which object was handled first, what remains inside a container, or where an object was placed before becoming occluded. Such tasks therefore require decisions that depend on interaction history under partial observability~\cite{robomme,memoryvla}. Yet retaining history alone is insufficient. A previously valid relation may cease to hold after a slip, human intervention, or environmental rearrangement. A robot must therefore preserve \emph{what happened} while revising \emph{what is currently supported by the available evidence}.

Generalist robot policies have substantially improved the execution of individual manipulation skills~\cite{rt2,octo,openvla,pi05,g05}. Recent models further extend language-conditioned control to long-horizon settings; in particular, G0.5 provides an autoregressive interface that interleaves reasoning and action generation within a single policy~\cite{g05}. However, a capable executor does not by itself determine what should happen next when the relevant evidence is old, hidden, or contradicted by a subsequent physical intervention.

Recent work equips manipulation systems with temporal context through learned memory banks, sparse visual events, or retrieved keyframes ~\cite{memoryvla,eventvla,memer}. Hierarchical systems instead place memory and reasoning above the motor policy: HiMe ~\cite{hime} maintains editable multimodal memory, while HyMeS ~\cite{hymes} and BATON \cite{baton} steer reusable or frozen skills using external verification and transition-aware memory. These methods establish that memory can extend a policy beyond its immediate observation. They also sharpen a different question: after evidence changes, which earlier conclusions should change with it?

\begin{figure}[!t]
\centering
\includegraphics[width=1\columnwidth]{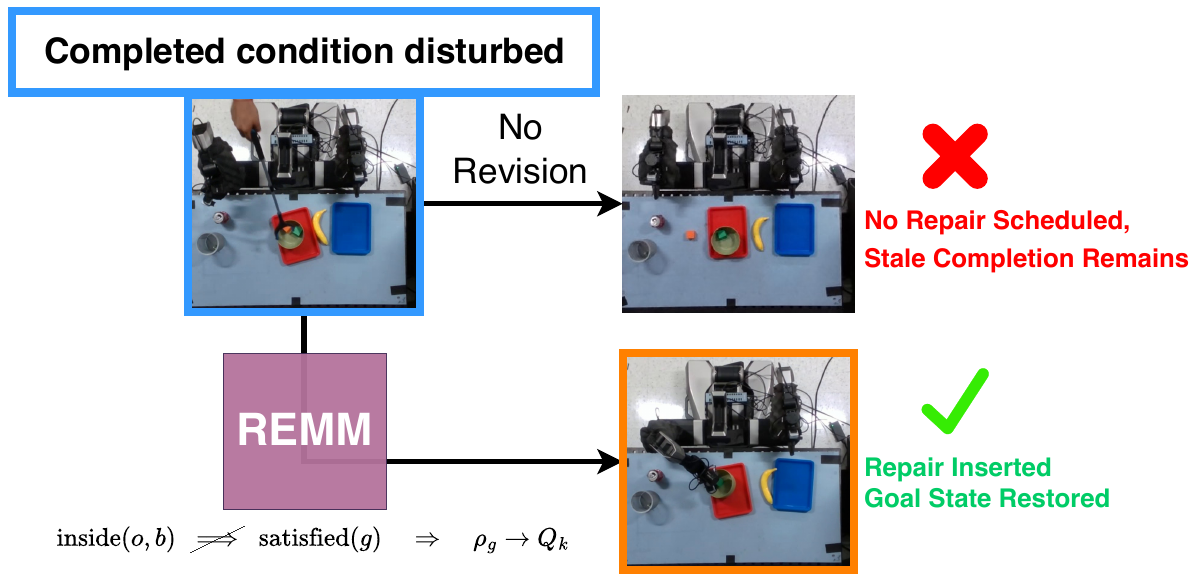}
\caption{\textbf{Revising memory after physical intervention.} A manipulation history can support either continued execution or recovery depending on whether later evidence preserves or contradicts an earlier task relation. \method updates the supported task state and changes the subsequent robot action without retraining the frozen executor.}
\label{fig:teaser}
\end{figure}

Closed-loop planners can incorporate success, scene, or human feedback into subsequent decisions~\cite{innermonologue}; structured world representations encode relational scene state~\cite{conceptgraphs}; and dynamic memories can add, delete, or modify stored relations as new evidence arrives~\cite{robomemory}. Yet these capabilities do not explicitly track which physical relations support previously completed task conditions. If such a relation changes without revising its downstream consequences, the controller can retain an internally coherent but physically stale task state. We therefore consider an intervention-grounded process memory in which physical relations, derived scene state, and completed task conditions are connected through explicit support dependencies. When verified evidence changes an upstream relation, its affected consequences can be selectively withdrawn. These dependencies are supplied by task and physical semantics rather than learned through causal discovery; their use as recorded justifications is related to truth maintenance~\cite{doyle1979}, while remaining distinct from causal world models that learn or probabilistically infer manipulation outcomes~\cite{causalworld,cobrapmm}.

We introduce \textbf{\method}, \textbf{R}evisable \textbf{E}mbodied \textbf{M}emory for \textbf{M}anipulation, a supervisory layer around a frozen G0.5 executor. \method preserves an append-only evidence history while maintaining a separate projection of currently supported state. It records causal dependency links from physical relations to derived and completed task conditions; when verified evidence supersedes a parent relation, \method invalidates affected descendants, identifies the broken task obligation, and schedules a bounded recovery instruction at the next verified stationary skill boundary. This produces an inspectable path from physical intervention to memory revision and changed control without \method-specific parameter training.

We evaluate \method on real-robot tasks involving transport disruptions, delayed sequence recall, and hidden-object rearrangement. The evaluation separates \emph{retention} from \emph{revision}: sequence and hidden-object tasks test whether decision-relevant history persists through delay and occlusion, while transport interventions test whether previously valid task state is correctly withdrawn and repaired after the world changes. Beyond end-to-end task success, we measure change acceptance, dependency revision, recovery timing and destination, physical repair, and the observed memory-to-control trace. Across 51 physical trials, including operator-assisted runs, \method improves mean recorded task progress over the strongest baseline from 57\% to 91\% in intervention recovery and from 43\% to 70\% in sequence recall, while completing all four of its hidden-object trials.\\

\noindent Our contributions are threefold:
\begin{itemize}
    \item a formulation of \emph{revisable embodied memory} that separates recorded execution history from currently supported task state and represents intervention-sensitive dependencies from physical relations to derived and completed task conditions;
    \item \method, a supervisory system that propagates dependency invalidation and dispatches nominal or recovery instructions to a frozen G0.5 policy without \method-specific parameter training; and
    \item a controlled real-robot evaluation that traces observed changes through memory revision to subsequent control decisions across transport, sequence-recall, and hidden-object manipulation tasks.
\end{itemize}

\begin{figure*}[!t]
\centering
\includegraphics[width=\textwidth]{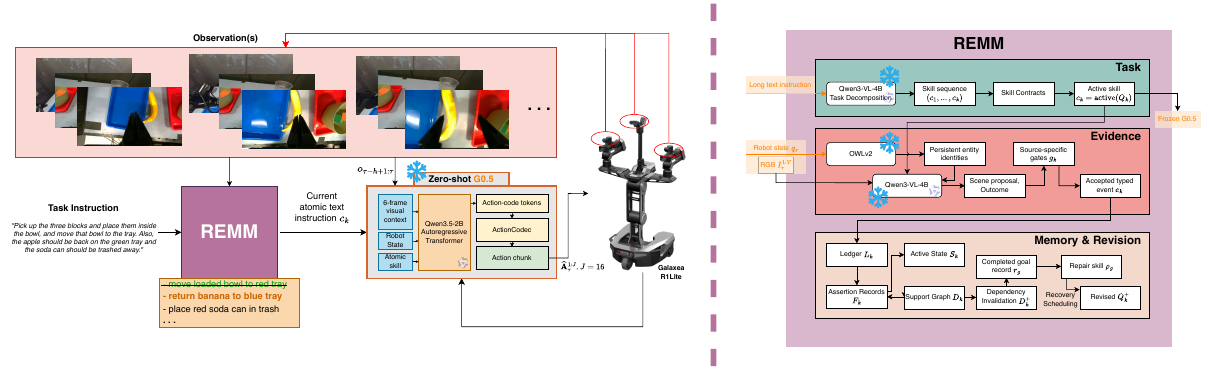}

\caption{\textbf{System and memory architecture.} Left: \method converts a long-horizon instruction and multi-view observations into the current atomic instruction for a frozen G0.5 executor, which predicts action-code tokens and action chunks for the Galaxea R1 Lite; new observations and robot state close the loop after execution. Right: a task decomposer and skill contracts define the active skill, frozen perception and visual verification produce accepted typed events, and an append-only ledger maintains an active-state projection and support graph. When new evidence supersedes a supporting fact, dependency invalidation withdraws affected goal records before \method schedules a repair skill or issues the revised next query. Snowflakes denote frozen models.}
\label{fig:architecture}
\end{figure*}

\section{Related Work}
\label{sec:related}

\subsection{Memory for Vision--Language--Action Control}

Policy-integrated approaches make history part of action generation. MemoryVLA retrieves perceptual and cognitive tokens from a learned memory bank~\cite{memoryvla}; EventVLA learns when to preserve sparse visual evidence~\cite{eventvla}; and Chameleon and ChainVLA propagate compact event or execution state across action queries~\cite{chameleon,chainvla}. These methods learn representations for retention and retrieval inside the policy. RoboMME shows that their benefit depends strongly on whether a task requires temporal, spatial, object, or procedural memory~\cite{robomme}. \method uses a different boundary: it leaves G0.5 frozen, does not learn a memory-conditioned action model, and studies explicit revision of task state between bounded skills.

Hierarchical approaches place memory above the executor. MemER trains a high-level visual policy to select keyframes and predict subtasks for a separately trained $\pi_{0.5}$~\cite{memer}. HiMe divides control among an Executor, a working-memory Sentry, and a Planner that retrieves and edits multimodal memory~\cite{hime}. HyMeS uses executable memory-management heuristics and multimodal completion checks to steer a Markovian VLA~\cite{hymes}, while BATON uses a frozen VLA with subtask exploration and transition-aware memory~\cite{baton}. These systems are the closest architectural context for \method. Our question is narrower than general memory management: whether storing explicit parents for derived and completed task conditions allows a verified intervention to revise exactly the downstream obligations that no longer hold.

\subsection{Closed-Loop State Revision and Recovery}

Robotic planners have long used feedback to reconsider high-level action. Inner Monologue conditions an LLM planner on success detection, scene description, and human feedback~\cite{innermonologue}; $\mathcal P^3$ schedules emerging tasks using priorities, dependencies, interruption, and resumption~\cite{zhou2026mathcalp3versatileembodiedagents}; and PhyAgentOS couples persistent session state with verification and replanning~\cite{phyagentos}. These works establish closed-loop supervisory control, but their task scheduling or model-mediated replanning does not by itself expose which physical fact sustained a completed condition. \method connects these levels explicitly and preserves the active atomic skill while inserting, deferring, or retargeting a recovery at a skill boundary.

Structured memories provide the complementary world-state representation. ReMEmbR indexes spatially and temporally grounded experience for navigation~\cite{remembr}; ConceptGraphs builds open-vocabulary object--relation graphs for planning~\cite{conceptgraphs}; and RoboMemory revises relevant neighborhoods of temporal, spatial, semantic, and episodic memory~\cite{robomemory}. \method does not claim the first relational memory or editable knowledge graph. It adds an append-only evidence ledger, an active projection, and typed links from physical relations to derived scene and task conditions.

Those links draw on two related traditions. Truth-maintenance systems retain justifications and withdraw conclusions when their support becomes inactive~\cite{doyle1979}. Causal manipulation work instead uses controlled interventions or causal Bayesian models to reason about action effects~\cite{causalworld,cobrapmm}. \method does not discover a causal graph or predict general dynamics. It instantiates task-semantic causal dependencies whose operational meaning is intervention-sensitive: changing a required parent relation changes the validity of its descendant task condition and, through the controller, the selected continuation. The experiments therefore test both the stored revision and its downstream control effect rather than treating an edited memory record as sufficient evidence of recovery.

\section{Method}
\label{sec:method}

\subsection{Problem Setting and System Overview}

The robot receives episode instruction $\ell$ and observations $o_\tau=(I_\tau^{1:V},q_\tau)$, comprising synchronized images from $V$ views and proprioception. At each policy query $\tau$, the frozen G0.5 executor consumes its latest $h$ observations, \method-selected atomic instruction $c_k$, and R1 Lite embodiment identifier $e_{\mathrm{R1}}$:

\begin{equation}
\left(y_\tau,\widehat{\mathbf A}_{\tau}^{1:J}\right)
=\pi_{\mathrm{G0.5}}\!\left(
o_{\tau-h+1:\tau},c_k,e_{\mathrm{R1}};\theta_0
\right),\quad \theta_0\ \text{fixed}.
\label{eq:executor}
\end{equation}

The outputs are optional grounded reasoning tokens $y_\tau$ and a continuous action chunk $\widehat{\mathbf A}_{\tau}^{1:J}$. In our deployment, the client executes $J=16$ actions at $15$ Hz. \method holds $c_k$ fixed during the skill and selects a new instruction only at a verified boundary; the executor parameters $\theta_0$ remain frozen.

At boundary $k$, \method maintains $M_k=(L_k,F_k,D_k,Q_k)$: the append-only event ledger $L_k$, assertion records $F_k$, directed support graph $D_k$, and ordered skill queue $Q_k$. The currently dispatched instruction is $c_k=\operatorname{active}(Q_k)$.

Separate from the G0.5 executor, \method uses frozen Qwen3-VL-4B-Instruct~\cite{qwen3vl} for constrained task decomposition and visual verification, and OWLv2-base-patch16-ensemble~\cite{owlv2} for open-vocabulary 2D object proposals. Deterministic validators and source-specific evidence gates convert these outputs into accepted typed events; T2 and T3 use task-specific memory-query compilers rather than open-ended Qwen planning.

\subsection{Revisable Memory State}

Let $\varphi_i\equiv[p_i(\mathbf x_i)=v_i]$ be the grounded proposition in assertion record $r_i$, where $p_i$ is a predicate, $\mathbf x_i$ contains its entity arguments, and $v_i\in\{\mathrm{true},\mathrm{false},\mathrm{unresolved}\}$. Each record also stores confidence $\gamma_i$, evidence references $\mathcal E_i$, direct parents $\mathcal P_i$, and status $z_i\in\{\mathrm{active},\mathrm{superseded},\mathrm{invalidated}\}$:

\begin{equation}
\begin{aligned}
r_i&=(\varphi_i,\gamma_i,\mathcal E_i,\mathcal P_i,z_i),\\
S_k&=\{\varphi_i\mid r_i\in F_k,\ z_i=\mathrm{active}\}.
\end{aligned}
\label{eq:assertion-state}
\end{equation}

Each $r_j\in\mathcal P_i$ creates edge $r_j\rightarrow r_i$ in $D_k$. The record set $F_k$ preserves active and inactive assertions, while $S_k$ projects only the propositions \method currently treats as supported. These two views answer different questions. The ledger records that an object was placed in a bowl; the projection answers whether its containment is still supported now. A later intervention can change the second answer without erasing the first.

Each derived record has one conjunctive support set: every parent is required. The implementation does not maintain competing alternative justifications. Losing any required parent therefore withdraws the derived record until new accepted evidence supports it again. An invalidated record is no longer justified by the stored evidence; it is not automatically a new observation that the proposition is false.

A candidate event $e_k$ contains a proposed state change and its provenance. Binary gate $g_s$ applies the acceptance rule for source $s=\operatorname{src}(e_k)$; only accepted evidence enters the ledger:

\begin{equation}
L_{k+1}=
\begin{cases}
L_k\mathbin{\|}e_k,
& g_{\operatorname{src}(e_k)}(e_k)=1,\\
L_k, & \text{otherwise},
\end{cases}
\label{eq:evidence-gate}
\end{equation}

where $\mathbin{\|}$ denotes append. Visual relations require grounded identities, an allowed relation, noncontradictory evidence, confidence of at least $0.8$, and two distinct agreeing observations. Operator-verified outcomes and geometric contradictions use separate gates; a missing detection alone never retracts a fact. Confidence is a source-specific acceptance signal, not a calibrated probability of correctness.

This boundary keeps a model proposal separate from an accepted change. An object disappearing from one camera view may reflect occlusion rather than motion, and a policy's completion statement may precede a failed release. Neither is sufficient by itself to revise the task state. Accepted events retain their entity and frame references so that the controller's later decision can be traced back to the evidence that made it valid.

\subsection{Dependency Revision and Recovery}

\method separates direct physical assertions from their consequences. Let $o$ be an object, $c$ a container or support, $l$ a coarse task region, and $R\in\{inside,on\}$. If active records place $o$ relative to $c$ and $c$ at $l$, \method derives the supported location of $o$:
\begin{equation}
\begin{aligned}
r_s &: R(o,c)=\top,\qquad R\in\{inside,on\},\\
r_l &: at(c,l)=\top,\\
r_s,r_l &\Longrightarrow r_d:at(o,l)=\top,\\
\mathcal P_d&=\{r_s,r_l\},\quad
\gamma_d=\min(\gamma_s,\gamma_l).
\end{aligned}
\label{eq:supported-location}
\end{equation}
Here $l$ is a symbolic workspace region. The derived record stores both parents and inherits the weaker confidence, creating an explicit dependency that can later be withdrawn~\cite{liang2016what,tenorth2017knowrob}.

\begin{table*}[!t]
\centering
\caption{\textbf{Evaluation coverage by task.} Checks mark mechanisms tested by at least one predeclared condition. T2 combines nominal ordered recall with disturbed repeats; only the disturbed repeats test revision and recovery.}
\label{tab:coverage}
\small
\setlength{\tabcolsep}{4pt}
\renewcommand{\arraystretch}{1.18}
\begin{tabular*}{\textwidth}{@{\extracolsep{\fill}}llccccccc@{}}
\toprule
& & \multicolumn{3}{c}{Memory demands} & \multicolumn{4}{c}{Revision demands}\\
\cmidrule(lr){3-5}\cmidrule(l){6-9}
Task & Condition & Retention & Order & Identity & Revision & Invalidation & Recovery & Retargeting\\
\midrule
\rowcolor{transportrow} T1a & Queued disruption & \evalyes & \evalno & \evalno & \evalyes & \evalyes & \evalyes & \evalno\\
\rowcolor{transportrow} T1b & Destination shift after pouring & \evalyes & \evalno & \evalno & \evalyes & \evalyes & \evalyes & \evalyes\\
\rowcolor{sequencerow} T2 & Ordered recall + disturbed repeats & \evalyes & \evalyes & \evalno & \evalyes & \evalyes & \evalyes & \evalno\\
\rowcolor{hiddenrow} T3 & Hidden-object shuffle & \evalyes & \evalno & \evalyes & \evalno & \evalno & \evalno & \evalno\\
\bottomrule
\end{tabular*}
\end{table*}

For atomic placement goal $g$, $\operatorname{verify}_k(g)=1$ denotes an accepted physical postcondition at boundary $k$. \method represents completion as a derived record supported by the verified physical relation:

\begin{equation}
\begin{aligned}
&\operatorname{verify}_k(g)=1,
\quad R\in\{inside,on,at\},\\
&r_s:R(o,d)=\top\\
&\quad\Longrightarrow r_g:satisfied(g)=\top,\\
&\mathcal P_g\supseteq\{r_s\}.
\end{aligned}
\label{eq:goal-support}
\end{equation}

Goal completion therefore remains linked to the observed physical condition that justified it. This follows the general principle of semantic execution monitoring---checking expected action effects against observations~\cite{bouguerra2008monitoring}---while \method additionally records the supporting fact.

Edges in $D_k$ encode declared support relations, related to causal-link planning and task decomposition~\cite{young1994decomposition}; \method supplies these links from task semantics rather than learning them from trajectories. Let $D_k^{+}$ denote reachability through one or more support edges. When record $r_i$ becomes inactive, \method invalidates every active descendant $r_j$:

\begin{equation}
\begin{aligned}
z_i&\neq\mathrm{active},\quad
(r_i,r_j)\in D_k^{+},\\
z_j&=\mathrm{active}
\ \Longrightarrow
z_j\leftarrow\mathrm{invalidated}.
\end{aligned}
\label{eq:invalidation}
\end{equation}

As in truth maintenance, this operation preserves the evidence history while withdrawing only conclusions supported by the changed relation~\cite{doyle1979}. If invalidation reaches completed-goal record $r_g$, \method creates or updates repair skill $\rho_g$. Let $E_g=\operatorname{ent}(\rho_g)$ and $E_k=\operatorname{ent}(c_k)$ contain the persistent entities involved in the repair and active instruction. Their overlap determines the queue update:

\begin{equation}
Q_k^{+}=
\begin{cases}
\operatorname{insert}_{\mathrm{next}}(Q_k,\rho_g),
& E_g\cap E_k\neq\varnothing,\\
\operatorname{append}(Q_k,\rho_g), & \text{otherwise}.
\end{cases}
\label{eq:recovery-schedule}
\end{equation}

Here $Q_k^{+}$ is the post-revision queue. The operator $\operatorname{insert}_{\mathrm{next}}$ places the repair immediately after the active skill, whereas $\operatorname{append}$ defers it behind the remaining work. Both operations deduplicate equivalent repairs and may retarget a pending repair using newer evidence. The active $c_k$ is unchanged; the controller dispatches from $Q_k^{+}$ only at a stationary skill boundary.

For example, suppose a verified placement supports both $inside(o,b)$ and the completed goal of transporting $o$ with bowl $b$. If later evidence places $o$ outside the bowl, the earlier containment record is superseded, its inferred destination and completion lose support, and the transport obligation becomes pending again. The bowl's independently verified location and unrelated completed placements remain active. The repair is compiled from the current supported state, rather than replaying a stale instruction. If the required destination changes before dispatch, the pending repair can change with it.

\subsection{From Retained History to an Atomic Instruction}

Revision is only one use of the same memory boundary. In T2, the demonstrated cue order remains historical information even after the cues are covered. The controller must then bind a remembered cue to an object visible in the current scene and resolve the placement destination. Retaining the color sequence alone does not supply either binding. The task-specific compiler uses these records to form the next bounded placement instruction; the executor receives that instruction and its native observations, not the full event ledger.

T3 uses a different retained relation: the target object was covered by a particular cup. The selected cup must follow that persistent identity through rearrangement rather than being reidentified only by its initial left, middle, or right position. The offline implementation uses pixel-initialized MIL tracking for this purpose~\cite{babenko2009mil}; a task compiler converts the tracked cover's final position into the lift instruction. This specialized path is distinct from T1 dependency revision and from open-ended VLM planning. We therefore evaluate retention, grounding, and repair separately rather than attributing every successful memory query to the support graph.

\section{Experiments}
\label{sec:experiments}

We evaluate \method across 51 physical robot trials and 42 matched generated-video clips. Our experiments address three questions:
\begin{enumerate}
    \item \textbf{Revision and recovery:} When new physical evidence contradicts a previously satisfied task condition, can \method invalidate the dependent state and schedule the correct repair?
    \item \textbf{Retention and grounding:} Can \method preserve ordered and object-identity information through delay, occlusion, and rearrangement, then ground that history into the correct next instruction?
    \item \textbf{Control effect and cost:} With the G0.5 executor held fixed, how do these memory decisions affect task progress, end-to-end success, failure modes, and system cost relative to no-memory and hierarchical-memory baselines?
\end{enumerate}

\subsection{Task Families}

\begin{figure}[!t]
\centering
\includegraphics[width=\columnwidth]{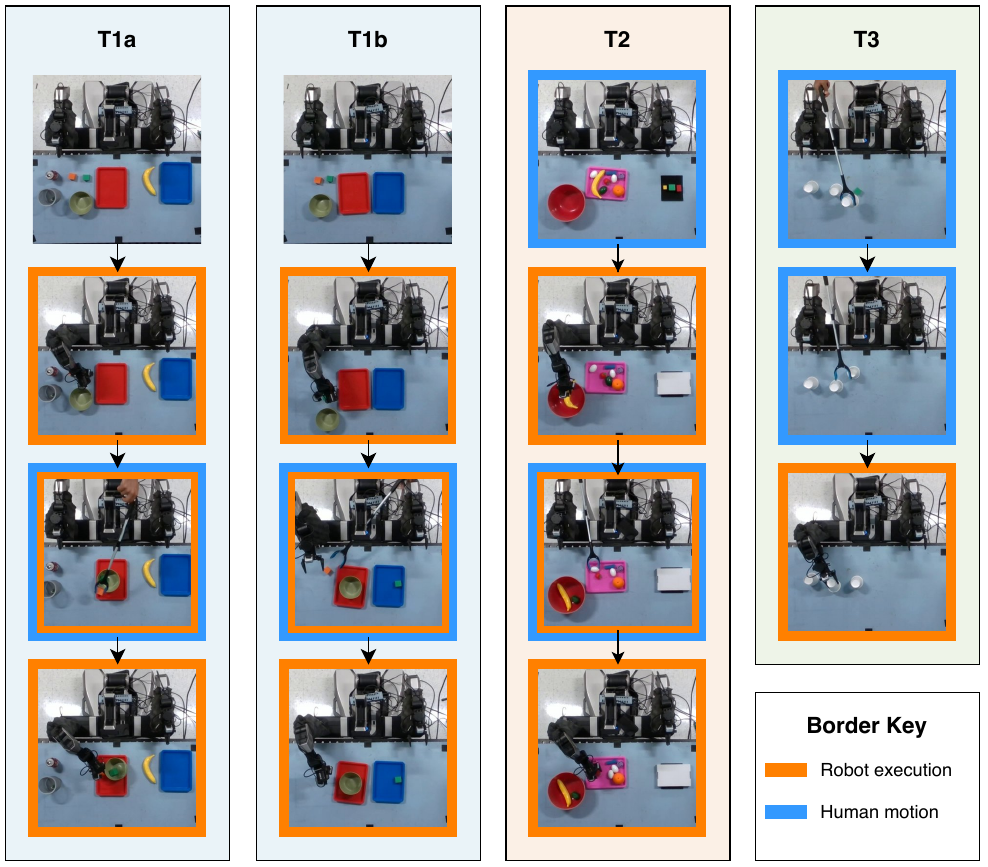}
\caption{\textbf{Physical task families.} T1a and T1b test recovery scheduling and retargeting; T2 tests ordered recall and disturbed repeats; T3 tests hidden-object identity. Blue borders mark human motion and orange borders robot execution; arrows show task progression.}
\label{fig:experiment-setup}
\end{figure}

We organize the study by memory demand, rather than by atomic motor skill, as summarized in Fig.~\ref{fig:experiment-setup}. 

\textbf{Transport and recovery} tests deferring a correction while unrelated work remains queued (T1a) and retargeting recovery after a destination change (T1b). 

\textbf{Sequence recall} (T2) tests recording a hidden demonstration order, grounding each historical cue to a current object, and compiling an ordered multi-skill queue; predeclared disturbed repeats additionally test whether completed placements are revised and repaired without corrupting the remaining order. 

\textbf{Hidden-object search} (T3) instead tests whether one object--cover binding remains attached to the same cover through a shuffle.

\subsection{Dependency Mechanism Controls}

To distinguish stored dependency invalidation from observing the same physical change, we define three state-update strategies. Let $\mathcal V_k$ contain completed-goal records newly invalidated at boundary $k$, let $h_k(r)=1$ when support record $r$ agrees with the latest direct facts, and let $b_k^{\sigma}(g)$ indicate that strategy $\sigma$ declares goal $g$ broken:
\begin{equation}
b_k^{\sigma}(g)=
\begin{cases}
\mathbb I[r_g\in\mathcal V_k], & \sigma=\mathrm{explicit},\\
\mathbb I[\exists r\in\mathcal P_g:h_k(r)=0],
& \sigma=\mathrm{recompute},\\
0, & \sigma=\mathrm{direct}.
\end{cases}
\label{eq:dependency-strategy}
\end{equation}
The explicit strategy reads \method's stored invalidation, recomputation checks the saved postconditions on demand, and direct update changes only the physical relation. All three receive the same accepted evidence; this is a software-level mechanism control, not an external baseline. The full-system physical comparisons below do not isolate an advantage of stored propagation over recomputing the affected goals.

\subsection{Evaluation Metrics}

For trial $e$, let $\mathcal G_e=(g_{e,1},\ldots,g_{e,K_e})$ be evaluator-defined task checkpoints fixed before execution, and let $y_{e,m}\in\{0,1\}$ indicate whether checkpoint $m$ is independently satisfied. The intended checkpoint measures are partial task progress $P_e$ and strict end-to-end success $S_e$:
\begin{equation}
\begin{aligned}
P_e&=\frac{1}{K_e}\sum_{m=1}^{K_e}y_{e,m},\\
S_e&=\prod_{m=1}^{K_e}y_{e,m}.
\end{aligned}
\label{eq:task-progress}
\end{equation}
Checkpoints are fixed before execution and shared across methods; they never use a controller's self-generated plan. The trial export uses recorded checkpoint fractions when available and completion-label or T3 decision-label fallbacks when progress is absent. We call this exported measure \emph{recorded task progress}; all reported progress percentages use it rather than implying complete independent checkpoint annotations for every trial. Full, partial, and failed outcomes retain their recorded labels rather than being reconstructed by thresholding progress or by applying $S_e$ to incomplete annotations. For disturbed trials, accepted change, invalidated task state, recovery placement in the queue, and physical repair are distinct events; an observed memory update is not itself physical success.
\begin{equation}
\begin{aligned}
B_w &= \sum_{i=1}^{N_w} |\mathrm{serialize}(W_i)|,\\
B_r &= \sum_{j=1}^{N_r} |\mathrm{serialize}(R_j)|,\\
\bar n_w &= N_w/N_q,\qquad \bar n_r=N_r/N_q,\\
\bar b_w &= B_w/N_q,\qquad \bar b_r=B_r/N_q.
\end{aligned}
\label{eq:memory-cost}
\end{equation}
Here $W_i$ and $R_j$ are method-native logical write and retrieval payloads, and $N_q$ counts fresh G0.5 action-chunk queries. Current observations, audit recordings, and model weights are reported separately rather than counted as memory traffic.

\section{Results}

We report three complementary views: physical outcome decomposition, task-level decision and cost profiles, and matched offline diagnostics of high-level memory readout. The physical export uses saved trial-ledger outcomes and operator annotations. Each rate uses the available eligible labels for its own denominator, and blank labels are excluded from that rate rather than silently treated as failures. Appendix~\ref{app:physical-accounting} gives the task counts and assistance accounting.

\subsection{End-to-End Physical Outcomes}

\begin{figure}[!t]
\centering
\includegraphics[width=\columnwidth]{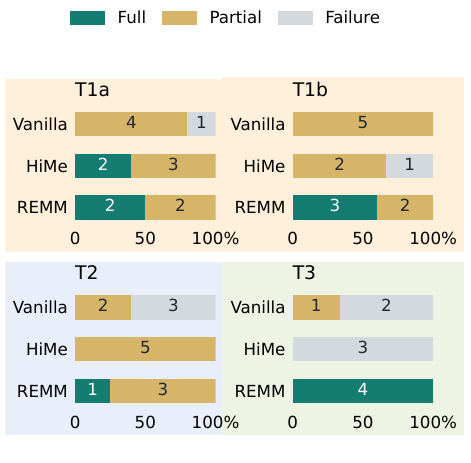}
\caption{\textbf{Physical outcomes by task.} Bars show full, partial, and failed outcomes with raw counts. Assisted runs remain included; HiMe denotes our adaptation around the shared G0.5 executor.}
\label{fig:outcome-decomposition}
\end{figure}

Figure~\ref{fig:outcome-decomposition} separates full task success from partial completion and failure. \method is the only system with a full completion in every task family: 10 of 17 trials overall, compared with 2 of 16 for HiMe and 0 of 18 for vanilla G0.5. None of the 17 \method trials is classified as a complete failure, although seven remain partial. These pooled counts are descriptive rather than a single comparable success rate because the task mix and denominators differ; the task-separated profiles below provide the more informative comparison.

These outcomes include operator-assisted runs. Two of \method's full completions have recorded task assistance, so the pooled counts are not unassisted-autonomy success rates. HiMe denotes our controlled adaptation around the shared G0.5 executor, rather than an official reproduction of the original model stack.

\begin{figure}[!t]
\centering
\includegraphics[width=\columnwidth]{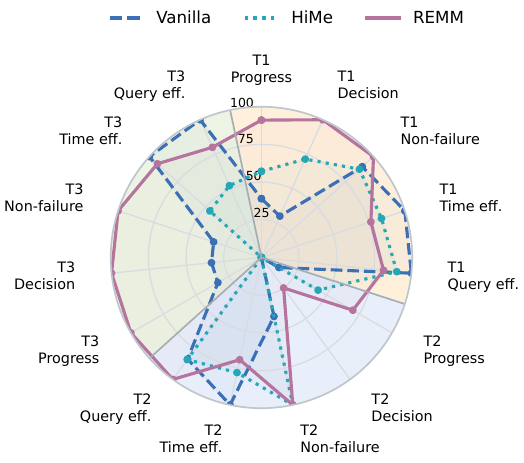}
\caption{\textbf{Task-level system profiles.} Task-colored sectors show recorded task progress, decision accuracy, non-failure completion, and relative time and G0.5-query efficiency. Progress includes the stated missing-data fallbacks. T1 pools T1a/T1b; higher is better. Non-failure includes partial trials; efficiency is $100\times$ best task-specific median cost / method median.}
\label{fig:system-profile}
\end{figure}

\subsection{Task-Level System Profiles}

Figure~\ref{fig:system-profile} reports recorded task progress, decision accuracy, non-failure completion, execution-time efficiency, and G0.5-query efficiency separately for T1, T2, and T3; T1 pools T1a and T1b. The first three measures are percentages from the available trial records, with the progress fallbacks defined above. For cost $x$, method $m$ on task $t$ receives efficiency $100\min_{m'}\widetilde{x}_{m',t}/\widetilde{x}_{m,t}$, where $\widetilde{x}_{m,t}$ is the median observed cost. This normalization preserves proportional cost differences while assigning 100 to the most efficient observed method per task.

The task-level profiles explain what the pooled outcomes hide. On T1, \method raises mean recorded task progress from 57\% for the strongest baseline to 91\%; on T2, it raises the same measure from 43\% to 70\%; and on T3 it completes all four trials while neither baseline records a full completion. This gain carries a visible systems cost: \method is not the most time- or query-efficient method on every task because it performs additional perception and verification and often continues runs that a baseline terminates early. We therefore report efficiency beside success rather than folding both into one score.

G0.5-query efficiency counts only fresh action-chunk queries, not planner, sentry, or verifier VLM calls. It is therefore not a measure of total inference cost. In T2, the baseline time and query medians each use only two of five trials. Missing records, assistance, and early stopping limit what can be concluded from these relative efficiencies; they do not establish lower cost at equal task performance.

\subsection{Offline Generated-Video Diagnostics}

The generated-video comparison uses the same 12 T2 clips and 30 T3 clips for every method; representative frames appear in Fig.~\ref{fig:video-frames}, and the complete result panels appear in Appendix~\ref{app:video-readout}, Fig.~\ref{fig:offline-score-bars}. T2 uses endpoint-conditioned LTX-Video 0.9.8 13B Distilled~\cite{ltxvideo,ltxmodels}, with OpenAI-generated endpoint images. T3 uses deterministic VET-Bench shell-game renders~\cite{vetbench}, rather than generative-model video. The scored snapshot stores human approval verdicts but not reviewer-count or adjudication records; a blinded second review remains needed for a fully documented benchmark release.

\begin{figure}[!t]
\centering
\includegraphics[width=.9\columnwidth]{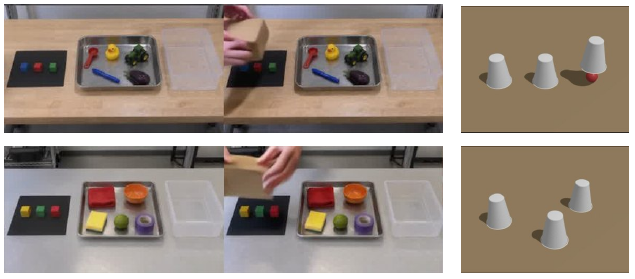}
\caption{\textbf{Offline video task families.} T2 shows cue demonstration and covering; T3 shows target covering and rearrangement. These diagnostics score high-level decisions without motor-policy execution.}
\label{fig:video-frames}
\end{figure}

T2 measures cue-order recall, exact three-item sequence, first-object and destination grounding, and a grounded first command; T3 requires one left, middle, or right cup choice and its corresponding lift command. The command scorer requires the correct action, object or color, and destination while allowing a fixed set of transparent noun equivalences. The no-memory condition uses final-frame Qwen3-VL-4B-Instruct, not the G0.5 motor policy. Adapted HiMe adds episodic and procedural memory. The MemER-style~\cite{memer} and RoboMME-style~\cite{robomme} conditions use frozen Qwen with visual-keyframe and symbolic-subgoal memory interfaces, respectively. They emit canonical language subtasks compatible with a downstream $\pi_{0.5}$ executor~\cite{pi05}; they are declared adaptations rather than official reproductions, and the passive-video diagnostic does not run the $\pi_{0.5}$ motor policy. No manifest-derived inventory, destination, expected command, or target answer is supplied to any method. Appendix~\ref{app:video-readout} describes what each output is compared against.

The three panels in Fig.~\ref{fig:offline-score-bars} separate remembering a sequence from turning that memory into an actionable instruction. MemER-style retains all 12 sequences but grounds neither the first object nor the command; RoboMME-style reaches 10/12 exact sequences, 6/12 first objects, and 6/12 commands. HiMe and \method both achieve cue LCS 35/36 and exact sequence 11/12, yet HiMe grounds the first object and command on only 1 of 12 clips, versus 11 of 12 and 8 of 12 for \method.

The five-metric T2 partial score is 86.1\% for \method, compared with 73.9\% for RoboMME-style, 55.0\% for MemER-style, 54.4\% for HiMe, and 38.9\% for no-memory. This descriptive average normalizes cue LCS by 36 and each remaining component by 12 before assigning equal weights. It is not the percentage of successful clips, and its two recall components are related measures rather than independent outcomes. On the stricter per-clip conjunction---correct sequence, first object, destination, and command---\method achieves 8/12 fully grounded T2 decisions, compared with 5/12 for RoboMME-style and 0/12 for MemER-style, HiMe, and no-memory. MemER-style's one-clip advantage in pure sequence retention and RoboMME-style's one-clip advantage in destination recognition therefore do not translate into a higher memory-to-action success rate; destination is visible in the final scene and is not itself the principal memory bottleneck.

In T3, the no-memory VLM always chooses the middle cup and therefore matches the fixed-set majority at 11/30; HiMe reaches 10/30, MemER-style 13/30, and RoboMME-style 11/30. \method's persistent target--cover track is correct on all 30 deterministic clips (Wilson 95\% interval: 88.6--100\%). This result is specific to the generated set and does not establish perfect tracking under real occlusion or appearance shift. These are high-level offline diagnostics rather than physical manipulation outcomes: passive video never executes G0.5, and observation selection and compute follow each method's architecture. In particular, \method uses eight sampled views with Qwen and event memory for T2, and a pixel-initialized MIL tracker with a task compiler for T3. These unequal observation and compute paths limit component-level conclusions.

\section{Discussion}
\label{sec:discussion}

\subsection{Interpreting the Results}

The results support separating retention, revision, and grounding rather than treating embodied memory as a single capability. Nominal T2 and T3 test whether task-relevant information remains available after delay, covering, or rearrangement. T1 and the disturbed T2 repeats additionally require a previously supported task condition to become inactive. \method is designed for this second case: accepted evidence is connected to the completed condition it supported, allowing a change to withdraw that condition and create a bounded repair instead of merely adding another observation to an undifferentiated history.

The offline T2 result clarifies where structured memory matters. MemER-style slightly exceeds \method on pure sequence storage, and RoboMME-style slightly exceeds it on visible destination recognition, but neither exceeds \method on the fully grounded decision. HiMe recalls the demonstrated order nearly as well as \method but rarely grounds that order to the correct first object or command. \method's advantage therefore lies in persistent entity binding and compilation of retained state into an atomic instruction, rather than universal dominance of every diagnostic column. The comparison remains system-level: each method uses its architecture-specific observation-selection and inference path, and the $\pi_{0.5}$-compatible conditions are evaluated only above the motor-policy boundary, so the result does not attribute the difference to one isolated component, memory size, or low-level executor.

Holding G0.5 fixed removes policy training as a confound and makes the selected high-level instruction auditable, but it does not make every end-to-end failure a memory failure. The methods can differ in when they query observations, when they stop, and which instruction they send; the common executor can still fail to realize a correct instruction. We therefore interpret decision correctness, recorded task progress, and physical success together. Likewise, the dependency controls isolate state-update logic in software, while the physical comparison measures complete systems; the physical gains should not be attributed to graph propagation alone.

\subsection{Scope and Limitations}

\method's dependency graph is supplied by the task compiler and a fixed predicate ontology. The recorded links have an intervention-sensitive causal interpretation within those declared task semantics: changing a required parent relation changes the validity of its dependent task condition. \method does not discover causal structure from trajectories, predict arbitrary physical effects, or automatically compile unconstrained language into new dependency rules.

The live state estimate also remains bounded by perception and execution. It uses conservative 2D detections, persistent identifiers, frozen VLM judgments, and source-specific evidence gates rather than calibrated metric 3D state. Missing detections do not retract facts, which reduces false revision but can delay necessary updates. Corrections are dispatched only at verified stationary skill boundaries; \method does not interrupt an unsafe action chunk in progress or guarantee recovery when identity is lost under complete occlusion.

Finally, the physical study uses one robot, a small tabletop object set, controlled interventions, and limited per-condition samples. HiMe is evaluated as a controlled adaptation around the shared executor, so the comparison characterizes these deployed systems rather than every configuration of the original method. The video diagnostics use generated T2 clips and deterministic T3 renders with architecture-specific compute paths. The MemER-style and RoboMME-style rows are likewise controlled, $\pi_{0.5}$-compatible high-level adaptations, not official reproductions or measurements of $\pi_{0.5}$ manipulation success. Broader evaluation should run the emitted subtasks through a matched low-level policy, test automatically constructed but auditable dependencies, calibrate 3D identity tracking, and study safe revision during execution across additional robots, policies, and task families. To support reproducibility, we will release our implementation, evaluation protocols, controller configurations, structured memory traces, and full trial videos through the project website.

\section{Conclusion}
\label{sec:conclusion}

Long-horizon manipulation requires a robot to retain useful history while withdrawing conclusions that later physical evidence no longer supports. \method addresses this problem with an append-only evidence ledger, a revisable task-state projection, and explicit task-semantic dependencies that connect physical relations to completed conditions and bounded recovery instructions. Across 51 physical trials, including operator-assisted runs, \method improves mean recorded task progress on intervention-recovery tasks from 57\% for the strongest baseline to 91\%. In matched video diagnostics, including $\pi_{0.5}$-compatible high-level planner adaptations, \method achieves 8/12 fully grounded T2 decisions versus 5/12 or fewer for every baseline, showing that retaining a sequence is insufficient unless it can be grounded into the next action. These results support revisable memory as a practical supervisory interface for a frozen VLA, while leaving open the harder problems of learning task dependencies, maintaining identity under unconstrained perception, and revising control safely within an active skill.

\bibliographystyle{IEEEtran}
\bibliography{references}

\begin{thebibliography}{10}
\providecommand{\url}[1]{#1}
\csname url@samestyle\endcsname
\providecommand{\newblock}{\relax}
\providecommand{\bibinfo}[2]{#2}
\providecommand{\BIBentrySTDinterwordspacing}{\spaceskip=0pt\relax}
\providecommand{\BIBentryALTinterwordstretchfactor}{4}
\providecommand{\BIBentryALTinterwordspacing}{\spaceskip=\fontdimen2\font plus
\BIBentryALTinterwordstretchfactor\fontdimen3\font minus
  \fontdimen4\font\relax}
\providecommand{\BIBforeignlanguage}[2]{{%
\expandafter\ifx\csname l@#1\endcsname\relax
\typeout{** WARNING: IEEEtran.bst: No hyphenation pattern has been}%
\typeout{** loaded for the language `#1'. Using the pattern for}%
\typeout{** the default language instead.}%
\else
\language=\csname l@#1\endcsname
\fi
#2}}
\providecommand{\BIBdecl}{\relax}
\BIBdecl

\bibitem{robomme}
\BIBentryALTinterwordspacing
Y.~Dai, H.~Fu, J.~Lee, Y.~Liu, H.~Zhang, J.~Yang, C.~Finn, N.~Fazeli, and
  J.~Chai, ``{RoboMME}: Benchmarking and understanding memory for robotic
  generalist policies,'' \emph{arXiv preprint arXiv:2603.04639}, 2026.
  [Online]. Available: \url{https://arxiv.org/abs/2603.04639}
\BIBentrySTDinterwordspacing

\bibitem{memoryvla}
H.~Shi, B.~Xie, Y.~Liu, L.~Sun, F.~Liu, T.~Wang, E.~Zhou, H.~Fan, X.~Zhang, and
  G.~Huang, ``Memoryvla: Perceptual-cognitive memory in vision-language-action
  models for robotic manipulation,'' \emph{arXiv preprint arXiv:2508.19236},
  2025.

\bibitem{rt2}
A.~Brohan \emph{et~al.}, ``{RT-2}: Vision-language-action models transfer web
  knowledge to robotic control,'' \emph{arXiv preprint arXiv:2307.15818}, 2023.

\bibitem{octo}
{Octo Model Team}, D.~Ghosh, H.~Walke, K.~Pertsch, K.~Black, O.~Mees
  \emph{et~al.}, ``Octo: An open-source generalist robot policy,'' in
  \emph{Robotics: Science and Systems}, 2024.

\bibitem{openvla}
M.~J. Kim, K.~Pertsch, S.~Karamcheti, T.~Xiao, A.~Balakrishna, S.~Nair,
  R.~Rafailov, E.~Foster, G.~Lam, P.~Sanketi, Q.~Vuong, T.~Kollar,
  B.~Burchfiel, R.~Tedrake, D.~Sadigh, S.~Levine, P.~Liang, and C.~Finn,
  ``{OpenVLA}: An open-source vision-language-action model,'' \emph{arXiv
  preprint arXiv:2406.09246}, 2024.

\bibitem{pi05}
{Physical Intelligence} \emph{et~al.}, ``$\pi_{0.5}$: A
  vision--language--action model with open-world generalization,'' \emph{arXiv
  preprint arXiv:2504.16054}, 2025.

\bibitem{g05}
\BIBentryALTinterwordspacing
Y.~Liu, Z.~Dong, B.~Ye, T.~Yuan, T.~Jiang, A.~Yang, S.~Cao, H.~Liu, Y.~Sun,
  Z.~Guo, X.~Liu, D.~Ke, C.~Pan, C.~Wu, T.~Cheng, X.~Ren, X.~Zhang, J.~Cui,
  Z.~Zhao, H.~Zhang, K.~Xu, H.~Yang, B.~Zhang, J.~Niu, S.~Zhu, S.~Zhang, and
  H.~Zhao, ``{G0.5}: One autoregressive stream for robot reasoning and
  action,'' 2026. [Online]. Available: \url{https://arxiv.org/abs/2608.11739}
\BIBentrySTDinterwordspacing

\bibitem{eventvla}
G.~Yang, Z.~Tu, Y.~Yang, S.~Mao, J.~Dong, T.~Chen, J.~Peng, J.~Xiong, J.~Cao,
  J.~Dai, W.~Zhou, Y.~Mu, and T.~Wang, ``{EventVLA}: Event-driven visual
  evidence memory for long-horizon vision-language-action policies,''
  \emph{arXiv preprint arXiv:2606.20092}, 2026.

\bibitem{memer}
A.~Sridhar, J.~Pan, S.~Sharma, and C.~Finn, ``{MemER}: Scaling up memory for
  robot control via experience retrieval,'' \emph{arXiv preprint
  arXiv:2510.20328}, 2025.

\bibitem{hime}
L.~Ji, S.~Wang, P.~Qian, X.~Yu, Y.~Tian, Z.~Fei, J.~Gong, and X.~Qiu, ``{HiMe}:
  Hierarchical embodied memory for long-horizon vision-language-action
  control,'' \emph{arXiv preprint arXiv:2607.03449}, 2026.

\bibitem{hymes}
Y.~Zhao, Z.~Ni, H.~Chen, R.~Zhang, and Q.~Zhu, ``Skills in weights, memory in
  code: Hybrid learning for memory-dependent robot manipulation,'' \emph{arXiv
  preprint arXiv:2608.09410}, 2026.

\bibitem{baton}
B.~Xu, Y.~Shang, and E.~Ferrara, ``Don't drop the {BATON}: Long-horizon robot
  manipulation via agentic subtask exploration and transition-aware memory,''
  \emph{arXiv preprint arXiv:2608.16889}, 2026.

\bibitem{innermonologue}
W.~Huang, F.~Xia, T.~Xiao, H.~Chan, J.~Liang, P.~Florence, A.~Zeng
  \emph{et~al.}, ``Inner monologue: Embodied reasoning through planning with
  language models,'' \emph{arXiv preprint arXiv:2207.05608}, 2022.

\bibitem{conceptgraphs}
Q.~Gu, A.~Kuwajerwala, S.~Morin, K.~M. Jatavallabhula, B.~Sen, A.~Agarwal,
  C.~Rivera, W.~Paul, K.~Ellis, R.~Chellappa, C.~Gan, C.~M. de~Melo, J.~B.
  Tenenbaum, A.~Torralba, F.~Shkurti, and L.~Paull, ``{ConceptGraphs}:
  Open-vocabulary 3d scene graphs for perception and planning,'' in \emph{2024
  IEEE International Conference on Robotics and Automation}, 2024, pp.
  5021--5028.

\bibitem{robomemory}
\BIBentryALTinterwordspacing
M.~Lei, H.~Cai, Y.~Yang, Y.~Wu, J.~Ren, Z.~Cui, L.~Tan, J.~Hong, G.~Hu, S.~Zhu,
  S.~Jiang, G.~Wang, J.~Tan, Z.~Wan, Z.~Li, Z.~Li, S.~Cui, Y.~Zhao, and Y.~Han,
  ``{RoboMemory}: A brain-inspired multi-memory agentic framework for
  interactive environmental learning in physical embodied systems,'' 2026.
  [Online]. Available: \url{https://arxiv.org/abs/2508.01415}
\BIBentrySTDinterwordspacing

\bibitem{doyle1979}
J.~Doyle, ``A truth maintenance system,'' \emph{Artificial Intelligence},
  vol.~12, no.~3, pp. 231--272, 1979.

\bibitem{causalworld}
O.~Ahmed, F.~Tr{\"a}uble, A.~Goyal, A.~Neitz, Y.~Bengio, B.~Sch{\"o}lkopf,
  M.~W{\"u}thrich, and S.~Bauer, ``{CausalWorld}: A robotic manipulation
  benchmark for causal structure and transfer learning,'' in
  \emph{International Conference on Learning Representations}, 2021.

\bibitem{cobrapmm}
R.~Cannizzaro, M.~Groom, J.~Routley, R.~O. Ness, and L.~Kunze, ``{COBRA-PPM}: A
  causal bayesian reasoning architecture using probabilistic programming for
  robot manipulation under uncertainty,'' in \emph{2025 European Conference on
  Mobile Robots}, 2025, pp. 1--8.

\bibitem{chameleon}
X.~Guo, C.~Jiang, H.~B. Kim, Y.~Han, Y.~Sun, Y.~Xiao, and J.~Yang, ``Chameleon:
  Control-indexed prospective memory for visuomotor manipulation,'' \emph{arXiv
  preprint arXiv:2603.24576}, 2026.

\bibitem{chainvla}
Y.~Huang, W.~Bu, Z.~Xiong, J.~Wu, F.~Huang, J.~Jiang, and Z.~Wang,
  ``{ChainVLA}: Chaining vision-language-action queries through a unified
  execution state for long-horizon manipulation,'' \emph{arXiv preprint
  arXiv:2608.02326}, 2026.

\bibitem{zhou2026mathcalp3versatileembodiedagents}
\BIBentryALTinterwordspacing
S.~Zhou, X.~Wang, J.~Zhang, R.~Tian, J.~Huang, R.~Xu, G.~Chen, and F.~Zheng,
  ``$\mathcal{P}^3$: Toward versatile embodied agents,'' 2026. [Online].
  Available: \url{https://arxiv.org/abs/2508.07033}
\BIBentrySTDinterwordspacing

\bibitem{phyagentos}
Y.~Liu, W.~Chen, X.~Song, T.~Pu, S.~Mo, Y.~Bai, Z.~Chen, Q.~Sun, L.~Zhong,
  Y.~Shen, and L.~Lin, ``{PhyAgentOS}: A self-evolving operating system for
  embodied agents with decoupled cognitive planning and physical execution,''
  \emph{arXiv preprint arXiv:2607.16636}, 2026.

\bibitem{remembr}
A.~Anwar, J.~Welsh, J.~Biswas, S.~Pouya, and Y.~Chang, ``{ReMEmbR}: Building
  and reasoning over long-horizon spatio-temporal memory for robot
  navigation,'' \emph{arXiv preprint arXiv:2409.13682}, 2024.

\bibitem{qwen3vl}
\BIBentryALTinterwordspacing
S.~Bai, Y.~Cai, R.~Chen \emph{et~al.}, ``{Qwen3-VL} technical report,''
  \emph{arXiv preprint arXiv:2511.21631}, 2025. [Online]. Available:
  \url{https://arxiv.org/abs/2511.21631}
\BIBentrySTDinterwordspacing

\bibitem{owlv2}
M.~Minderer, A.~Gritsenko, and N.~Houlsby, ``Scaling open-vocabulary object
  detection,'' in \emph{Advances in Neural Information Processing Systems},
  2023.

\bibitem{liang2016what}
W.~Liang, Y.~Zhao, Y.~Zhu, and S.-C. Zhu, ``What is where: Inferring
  containment relations from videos,'' in \emph{Proceedings of the Twenty-Fifth
  International Joint Conference on Artificial Intelligence}, 2016, pp.
  3418--3424.

\bibitem{tenorth2017knowrob}
M.~Tenorth and M.~Beetz, ``Representations for robot knowledge in the {KnowRob}
  framework,'' \emph{Artificial Intelligence}, vol. 247, pp. 151--169, 2017.

\bibitem{bouguerra2008monitoring}
A.~Bouguerra, L.~Karlsson, and A.~Saffiotti, ``Monitoring the execution of
  robot plans using semantic knowledge,'' \emph{Robotics and Autonomous
  Systems}, vol.~56, no.~11, pp. 942--954, 2008.

\bibitem{young1994decomposition}
R.~M. Young, M.~E. Pollack, and J.~D. Moore, ``Decomposition and causality in
  partial-order planning,'' in \emph{Proceedings of the Second International
  Conference on Artificial Intelligence Planning Systems}, 1994, pp. 188--193.

\bibitem{babenko2009mil}
B.~Babenko, M.-H. Yang, and S.~Belongie, ``Visual tracking with online multiple
  instance learning,'' in \emph{IEEE Conference on Computer Vision and Pattern
  Recognition}, 2009.

\bibitem{ltxvideo}
Y.~HaCohen, N.~Chiprut, B.~Brazowski, D.~Shalem, D.~Moshe, E.~Richardson,
  E.~Levin, G.~Shiran, N.~Zabari, O.~Gordon, P.~Panet, S.~Weissbuch,
  V.~Kulikov, Y.~Bitterman, Z.~Melumian, and O.~Bibi, ``{LTX-Video}: Realtime
  video latent diffusion,'' \emph{arXiv preprint arXiv:2501.00103}, 2024.

\bibitem{ltxmodels}
\BIBentryALTinterwordspacing
{Lightricks}, ``{LTX-Video}: Official model card and checkpoints,'' n.d., 13B
  0.9.8 distilled checkpoint; accessed September 24, 2026. [Online]. Available:
  \url{https://huggingface.co/Lightricks/LTX-Video}
\BIBentrySTDinterwordspacing

\bibitem{vetbench}
T.~Liu and W.~S. Lee, ``Can vision-language models solve the shell game?''
  \emph{arXiv preprint arXiv:2603.08436}, 2026.

\end{thebibliography}

\appendices
\section{Memory Records and Update Semantics}
\label{app:memory}

\begin{figure*}[!t]
    \centering
    \begin{minipage}[c]{0.52\textwidth}
    \includegraphics[width=\linewidth,trim=0 0 0 160,clip]{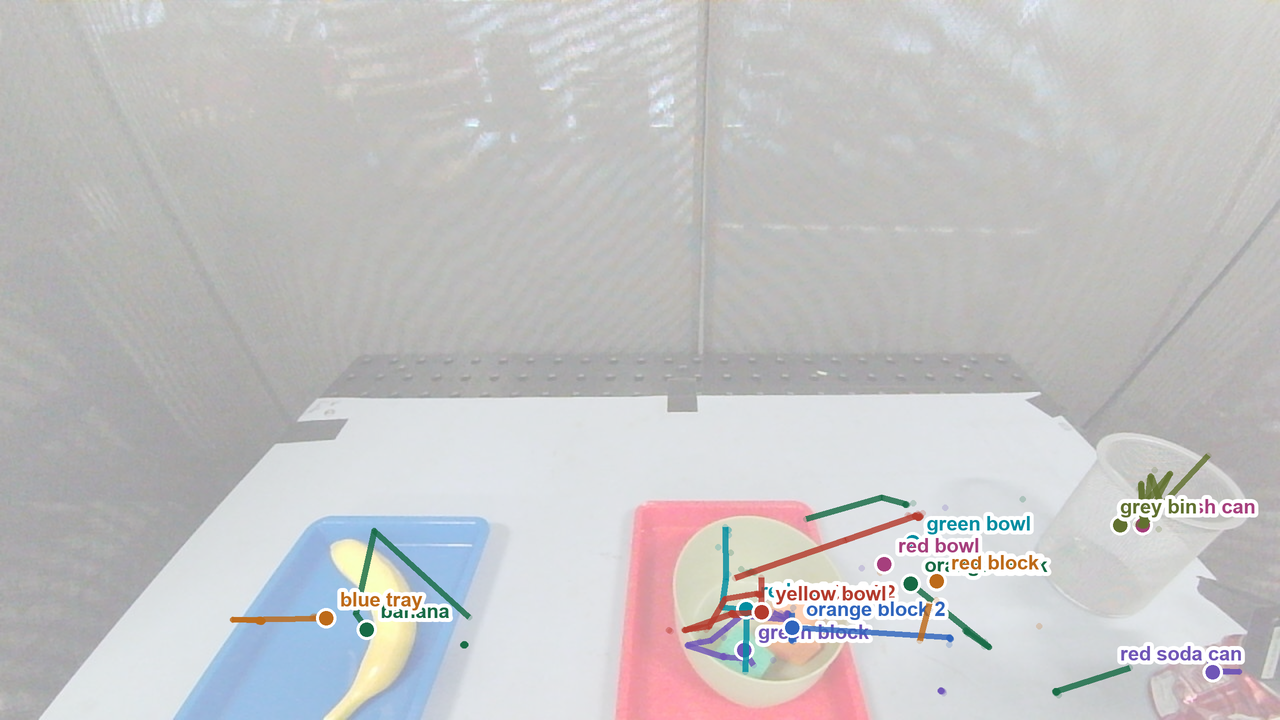}
    \end{minipage}\hfill
    \fbox{\begin{minipage}[c]{0.43\textwidth}
        \raggedright
        \fontsize{7.5}{9.5}\selectfont\ttfamily
        \{ "event\_id": "...\_56\_orange\_block\_01",\par
        \hspace*{0.7em}"evidence\_refs": [\par
        \hspace*{1.4em}"000056\_head",\par
        \hspace*{1.4em}"000056\_left\_wrist",\par
        \hspace*{1.4em}"000056\_right\_wrist"],\par
        \hspace*{0.7em}"fact\_id": "...\_56\_orange\_block\_01\_fact\_01",\par
        \hspace*{0.7em}"predicate": "inside", "value": false,\par
        \hspace*{0.7em}"status": "active",\par
        \hspace*{0.7em}"supersedes\_fact\_ids": ["...\_fact\_02"],\par
        \hspace*{0.7em}"recovery\_history": \{\par
        \hspace*{1.4em}"event\_id": "...\_56\_orange\_block\_01" \},\par
        \hspace*{0.7em}"decision": "recover\_affected\_object" \}\par
    \end{minipage}}
    \caption{\textbf{Recorded T1b memory example.} The trace contains 2,408 head-camera positions for 15 identities; gaps are disconnected tracks. The joined excerpt exposes evidence-to-event, fact-supersession, and recovery references. This assisted partial rollout uses the recompute-on-demand control and is included only for mechanism inspection.}
    \label{fig:memory-trace-json}
\end{figure*}

\subsection{Stored Records and References}
Accepted observations become immutable events containing entity identifiers, predicate changes, confidence values, and frame references. Events create revisable facts, while subgoal records retain completion and recovery evidence. The \texttt{supersedes\_fact\_ids} and \texttt{depends\_on\_fact\_ids} fields trace each update from its visual evidence to the affected goal and repair.

\subsection{Boundary Updates and Recovery}
A visual relation is accepted when identities are grounded, confidence reaches $0.8$, and two distinct observations agree; missing detections alone do not create negative facts. At stationary boundaries, accepted events supersede conflicting records, invalidate unsupported descendants, and expose affected task obligations. The controller then compiles or updates a deduplicated repair using the currently supported destination. A completed condition losing support does not authorize an arbitrary new plan: the repair remains bounded by the affected skill contract.

Figure~\ref{fig:memory-trace-json} joins evidence, supersession, and recovery references from an assisted partial T1b rollout using the recompute-on-demand control. It illustrates how those references can be inspected, not a successful explicit-invalidation ablation. The raw history and the interpretation of that history must remain distinguishable when inspecting any rollout.

\subsection{A Worked Support-Withdrawal Example}
Consider an accepted containment record $r_s$ and a separately accepted bowl-location record $r_l$. Their conjunction supports an inferred object location $r_d$, and a verified placement supports completion $r_g$. If a new event supersedes $r_s$, the explicit strategy traverses its outgoing support links and withdraws active descendants. The ledger still contains the placement event and the later intervention; $r_l$ remains active because it did not depend on the withdrawn containment. The object location and completion must now be established again before the transport goal can be treated as satisfied.

This example also explains the difference between negative and unsupported state. An accepted observation that the object is outside the bowl can create an active negative containment fact. The previous derived destination is merely invalidated: loss of containment does not by itself tell the controller where the object landed. Recovery therefore needs current grounding rather than interpreting every invalidated proposition as its opposite.

\section{Physical Evaluation and Accounting}
\label{app:physical-accounting}

\subsection{Shared Executor and Baseline Boundary}
The physical comparison uses Galaxea R1 Lite with shared cameras and a frozen G0.5 executor. All methods dispatch language instructions through the same motor-policy interface. Our HiMe adaptation preserves planner and sentry roles, episodic and procedural stores, and memory addition, update, and deletion~\cite{hime}. Its configured planner and sentry use Qwen3-VL-4B-Instruct, MiniLM-L6-v2 retrieval, and an eight-frame working context, with prompts and instruction parsing adapted to G0.5. These differences from the original model stack are why we refer to it as an adaptation.

Editable memory is not unique to \method. The more specific distinction is that \method retains the supporting records for a completed condition and connects their withdrawal to a repair obligation. Comparisons between complete systems also include perception, verification, instruction selection, and stopping behavior. They do not isolate graph propagation as the only changed component.

\begin{table}[t]
\centering
\caption{\textbf{Physical outcome counts.} Entries are full / partial / failure; their sum is the method's task denominator. Assisted runs remain included.}
\label{tab:physical-counts}
\small
\begin{tabular}{@{}lccc@{}}
\toprule
Task & Vanilla G0.5 & HiMe adaptation & \method\\
\midrule
T1a & 0 / 4 / 1 & 2 / 3 / 0 & 2 / 2 / 0\\
T1b & 0 / 5 / 0 & 0 / 2 / 1 & 3 / 2 / 0\\
T2 & 0 / 2 / 3 & 0 / 5 / 0 & 1 / 3 / 0\\
T3 & 0 / 1 / 2 & 0 / 0 / 3 & 4 / 0 / 0\\
\midrule
Total & 0 / 12 / 6 & 2 / 10 / 4 & 10 / 7 / 0\\
\bottomrule
\end{tabular}
\end{table}

\subsection{Assistance and Outcome Labels}
Table~\ref{tab:physical-counts} reproduces the task-separated outcome ledger. Assistance is recorded separately from outcome; the export does not relabel a completed run as autonomous simply because it reached the goal. The assistance fields contain 12 recorded-none, 0 task-assisted, and 6 unrecorded vanilla trials; 6, 2, and 8 HiMe trials; and 10, 4, and 3 \method trials, respectively. Unrecorded assistance does not mean unassisted.

The two task-assisted \method full completions are a T1b run with an orange-block nudge and a T2 run with help completing strawberry pickup. We retain the recorded outcome labels while making this assistance visible. Controller stop, return-home motion, and a policy's completion claim are not independent evidence of physical success. The saved ledger and operator annotations support the plotted export; a complete independent re-adjudication is not established by these figures.

\subsection{Reading the Radial Profile}
The radial plot groups five measures within each task sector. Recorded task progress uses the checkpoint fraction when available, with the stated completion-label or T3 decision-label fallbacks otherwise. Decision accuracy uses available high-level-choice labels: recovery timing or destination for T1, complete order for T2, and target-cover choice for T3. Non-failure completion includes both full and partial outcomes. A run can therefore have a correct decision or be non-failed without completing the task.

The labeled decision denominators are 10/8/9 for vanilla/HiMe/\method in pooled T1, 3/3/4 in T2, and 3/3/4 in T3. Their correct-decision counts are 3/5/9, 0/0/1, and 1/0/4, respectively. Each metric excludes its own blank labels, so these denominators need not match the outcome totals.

Time and query efficiency are relative median-cost scores, not elapsed seconds or query counts. Each task's lowest observed median receives 100; a method with twice that median would receive 50. T2 time and query medians use two of five baseline trials and all four \method trials. This available-record comparison is useful for inspecting overhead, but not for concluding that a system is cheaper at the same level of task completion. Supervisory VLM calls are excluded from the G0.5-query axis, and the plotted export does not establish total model-inference cost or complete memory-traffic accounting.

\section{Video Baseline Interfaces and Scoring}
\label{app:video-readout}

\begin{figure*}[!t]
\centering
\includegraphics[width=\textwidth]{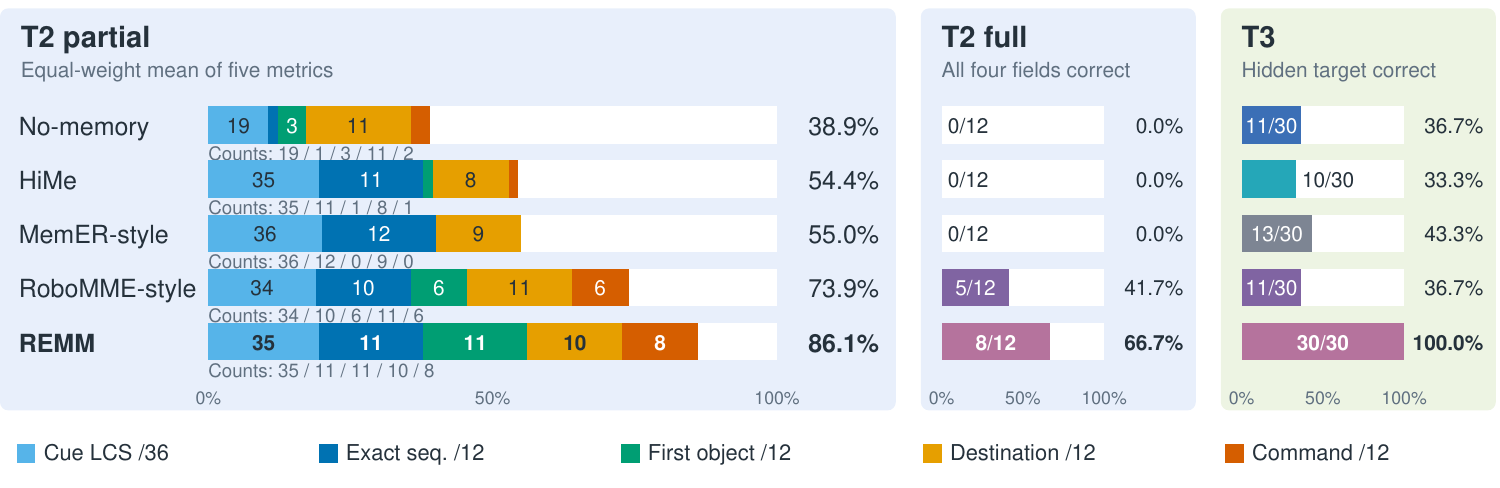}
\caption{\textbf{Offline video results.} T2 partial averages five normalized accuracies with equal weights; segment labels give raw counts in legend order. Full T2 requires correct sequence, first object, destination, and command on the same clip (/12). T3 scores hidden-target choice (/30). No motor policy is executed.}
\label{fig:offline-score-bars}
\end{figure*}

\subsection{What Each Baseline Predicts}
The offline methods produce a high-level readout, not action chunks. For T2, the scored output contains a predicted cue sequence, a first-object grounding, a destination, and a first language command. For T3, the output selects a final left, middle, or right cover position with its lift command. These are the fields consumed by the scorer; formatting a language subtask for a downstream executor does not demonstrate that the executor would carry it out.

The no-memory VLM reads the final frame without a retained demonstration. HiMe adds retrieved episodic and procedural memory. The MemER-style adaptation exposes retained visual keyframes to a high-level planner; the RoboMME-style adaptation exposes symbolic subgoal memory. Both use frozen Qwen in this comparison rather than the original trained model stack. The terms MemER-style and RoboMME-style identify those memory-interface choices, not official reproductions of MemER or RoboMME's policy variants~\cite{memer,robomme}.

The resulting language subtasks are compatible with a downstream $\pi_{0.5}$ interface~\cite{pi05}, but no $\pi_{0.5}$ motor inference or physical rollout occurs. \method's T2 path instead joins sampled visual evidence to event memory and a task-specific grounding compiler; its T3 path follows the initially bound target cover using MIL tracking~\cite{babenko2009mil}. Observation selection and compute differ across these paths. The matched clips control the diagnostic input set, not every aspect of the inference budget.

\subsection{Prediction Versus Expected Answer}
For each T2 clip, the scorer retains the demonstrated three-cue sequence and the expected first object, destination, and command. Cue LCS is the length of the longest common subsequence between predicted and demonstrated cues. Exact sequence requires the entire three-item order to match. First-object correctness tests the selected current object corresponding to the first remembered cue; destination correctness tests the required receiving object or region. Command correctness checks action, object or color, and destination together using the fixed noun-equivalence rules. A correct sequence does not automatically receive first-object or command credit.

Across 12 clips, three cue positions per clip give a maximum LCS total of 36. Each other T2 component is binary per clip and therefore has a denominator of 12. For example, the MemER-style row's 36 LCS and 12 exact sequences indicate complete sequence recall on this set. Its zero first-object and command counts are separate scoring failures, not contradictions of those recall scores. Reporting both makes the memory-to-action gap visible without altering any result.

For T3, the expected answer is the final position of the cover bound to the target, not its original position before the shuffle. We report one hidden-target score because the target selection and corresponding lift command represent the same decision in this diagnostic. The scorer's answer and generation metadata remain withheld from the methods; the reported results cover only the 12 T2 and 30 T3 clips in this snapshot.

\begin{samepage}
\subsection{Partial Accuracy and Full T2}
Let $a_m^{\mathrm{LCS}}$ be method $m$'s cue LCS total and let $a_m^{\mathrm{seq}},a_m^{\mathrm{obj}},a_m^{\mathrm{dest}},a_m^{\mathrm{cmd}}$ be its four binary-field totals. The plotted composite is
\begin{equation}
\begin{aligned}
A_m^{\mathrm{partial}}=\frac{100}{5}\Bigg(&\frac{a_m^{\mathrm{LCS}}}{36}
+\frac{a_m^{\mathrm{seq}}}{12}+\frac{a_m^{\mathrm{obj}}}{12}\\
&+\frac{a_m^{\mathrm{dest}}}{12}+\frac{a_m^{\mathrm{cmd}}}{12}\Bigg).
\end{aligned}
\label{eq:video-partial}
\end{equation}
\end{samepage}
Each colored segment in Fig.~\ref{fig:offline-score-bars} contributes one fifth of its normalized metric accuracy. Normalization is important: summing the raw counts would give cue LCS three times the available points of any other component. The chosen average gives the five displayed metrics equal weight, while still giving recall two related components. It is a descriptive summary, not five independent trials or a probability of motor success.

Full T2 instead counts clips on which exact sequence, first object, destination, and command are all correct together. It cannot be reconstructed by averaging those marginal totals: a method can get different fields right on different clips. \method's 8/12 full decisions and RoboMME-style's 5/12 therefore describe a stricter outcome than their respective 86.1\% and 73.9\% partial scores. The full T2 and T3 panels remain separate from physical-trial success.

\section{Scope of the Mechanism Evidence}
\label{app:mechanism-scope}
The direct-update, recompute-on-demand, and explicit-invalidation strategies distinguish three ways of responding to the same accepted change. Direct update revises the physical record without reopening its dependent goal. Recompute-on-demand checks saved postconditions when deciding what to do next. Explicit invalidation follows stored support links and exposes unsupported goals as soon as their required parents become inactive. This distinction concerns state-update logic, not which strategy can physically grasp an object.

The worked example and recorded references make the revision path inspectable, but the reported physical outcomes compare complete deployed systems. Establishing a separate benefit of support-graph propagation over recomputation would require fixed perception, accepted evidence, goal checks, and execution, followed by matched rollout comparisons. Likewise, nominal T2 and deterministic T3 primarily test retention and grounding. We do not treat their scores as evidence that dependency propagation alone caused the observed gains.

\end{document}